\documentclass{article}

\usepackage{microtype}
\usepackage{graphicx}
\usepackage{subcaption}
\usepackage{booktabs} 
\usepackage{amsmath}
\usepackage{amssymb}

\usepackage[accepted]{icml2018}

\icmltitlerunning{Neural Processes}

\begin{document}

\twocolumn[
\icmltitle{Neural Processes}




\begin{icmlauthorlist}
\icmlauthor{Marta Garnelo}{goo}
\icmlauthor{Jonathan Schwarz}{goo}
\icmlauthor{Dan Rosenbaum}{goo}
\icmlauthor{Fabio Viola}{goo}
\icmlauthor{Danilo J. Rezende}{goo}
\icmlauthor{S. M. Ali Eslami}{goo}
\icmlauthor{Yee Whye Teh}{goo}
\end{icmlauthorlist}

\icmlaffiliation{goo}{DeepMind, London, UK}

\icmlcorrespondingauthor{Marta Garnelo}{garnelo@google.com}

\icmlkeywords{Machine Learning, ICML}

\vskip 0.3in
]



\printAffiliationsAndNotice{} 

\newcommand{\marta}[1]{\textcolor{red}{\bf \small [#1 --marta]}}
\newcommand{\dan}[1]{\textcolor{blue}{\bf \small [#1 --dan]}}
\newcommand{\js}[1]{\textcolor{green}{\bf \small [#1 --js]}}
\newcommand{\yw}[1]{\textcolor{cyan}{\bf \small [#1 --yw]}}
\newcommand{\danilo}[1]{\textcolor{grey}{\bf \small [#1 --danilo]}}
\definecolor{purple}{RGB}{148, 0, 211}
\newcommand{\fabio}[1]{\textcolor{purple}{\bf \small [#1 --fabio]}}
\newcommand{\ali}[1]{\textcolor{red}{\bf \small [#1 --ali]}}
\def\EE{\mathbb{E}}

\begin{abstract}

A neural network (NN) is a parameterised function that can be tuned via gradient descent to approximate a labelled collection of data with high precision. A Gaussian process (GP), on the other hand, is a probabilistic model that defines a \textit{distribution} over possible functions, and is updated in light of data via the rules of probabilistic inference. GPs are probabilistic, data-efficient and flexible, however they are also computationally intensive and thus limited in their applicability. We introduce a class of neural latent variable models which we call \textit{Neural Processes} (NPs), combining the best of both worlds. Like GPs, NPs define distributions over functions, are capable of rapid adaptation to new observations, and can estimate the uncertainty in their predictions. Like NNs, NPs are computationally efficient during training and evaluation but also learn to adapt their priors to data. We demonstrate the performance of NPs on a range of learning tasks, including regression and optimisation, and compare and contrast with related models in the literature.

\end{abstract}
\section{Introduction}
\label{introduction}

Function approximation lies at the core of numerous problems in machine learning and one approach that has been exceptionally popular for this purpose over the past decade are deep neural networks.  
At a high level neural networks constitute black-box function approximators that learn to parameterise a single function from a large number of training data points. 
As such, the majority of the workload of a networks falls on the training phase while the evaluation and testing phases are reduced to quick forward-passes. 
Although high test-time performance is valuable for many real-world applications, the fact that network outputs cannot be updated after training may be undesirable. 
Meta-learning, for example, is an increasingly popular field of research that addresses exactly this limitation~\citep{sutskever2014sequence, wang2016learning, vinyals2016matching, finn2017model}. 

As an alternative to using neural networks one can also perform inference on a stochastic process in order to carry out function regression. The most common instantiation of this approach is a Gaussian process (GP), a model with complimentary properties to those of neural networks:
GPs do not require a costly training phase and can carry out inference about the underlying ground truth function conditioned on some observations, which renders them very flexible at test-time. In addition GPs represent infinitely many different functions at locations that have not been observed thereby capturing the uncertainty over their predictions given some observations.
However, GPs are computationally expensive: in their original formulation they scale cubically with respect to the number of data points, and current state of the art approximations still scale quadratically~\citep{quinonero2005unifying}.
Furthermore, the available kernels are usually restricted in their functional form and an additional optimisation procedure is required to identify the most suitable kernel, as well as its hyperparameters, for any given task. 

As a result, there is growing interest in combining aspects of neural networks and inference on stochastic processes as a potential solution to some of the downsides of both~\citep{huang2015scalable, wilson2016deep}.
In this work we introduce a neural network-based formulation that learns an approximation of a stochastic process, which we term \textit{Neural Processes} (NPs). 
NPs display some of the fundamental properties of GPs, namely they learn to model distributions over functions, are able to estimate the uncertainty over their predictions conditioned on context observations, and shift some of the workload from training to test time, which allows for model flexibility. 
Crucially, NPs generate predictions in a computationally efficient way.
Given $n$ context points and $m$ target points, inference with a trained NP corresponds to a forward pass in a deep NN, which scales with $\mathcal{O}(n+m)$ as opposed to the $\mathcal{O}((n+m)^3)$ runtime of classic GPs. 
Furthermore the model overcomes many functional design restrictions by learning an implicit ‘kernel’ from the data directly. 

\pagebreak
Our main contributions are:
\begin{enumerate}
\item We introduce Neural Processes, a class of models that combine benefits of neural networks and stochastic processes.
\item We compare NPs to related work in meta-learning, deep latent variable models and Gaussian processes. Given that NPs are linked to many of these areas, they form a bridge for comparison between many related topics.
\item We showcase the benefits and abilities of NPs by applying them to a range of tasks including 1-D regression, real-world image completion, Bayesian optimization and contextual bandits.
\end{enumerate}

\section{Model}
\label{model}

\subsection{Neural processes as stochastic processes}
\label{process}

The standard approach to defining a stochastic process is via its finite-dimensional marginal distributions. 
Specifically, we consider the process as a random function $F:\mathcal{X}\rightarrow \mathcal{Y}$ and for each finite sequence $x_{1:n}=(x_1,\ldots,x_n)$ with $x_i\in \mathcal{X}$, we define the marginal joint distribution over the function values $Y_{1:n}:=(F(x_1),\ldots,F(x_n))$. For example, in the case of GPs, these joint distributions are multivariate Gaussians parameterised by a mean and a covariance function.

\begin{figure*}
    \centering
    \begin{subfigure}[b]{0.29\textwidth}
        \includegraphics[width=\textwidth]{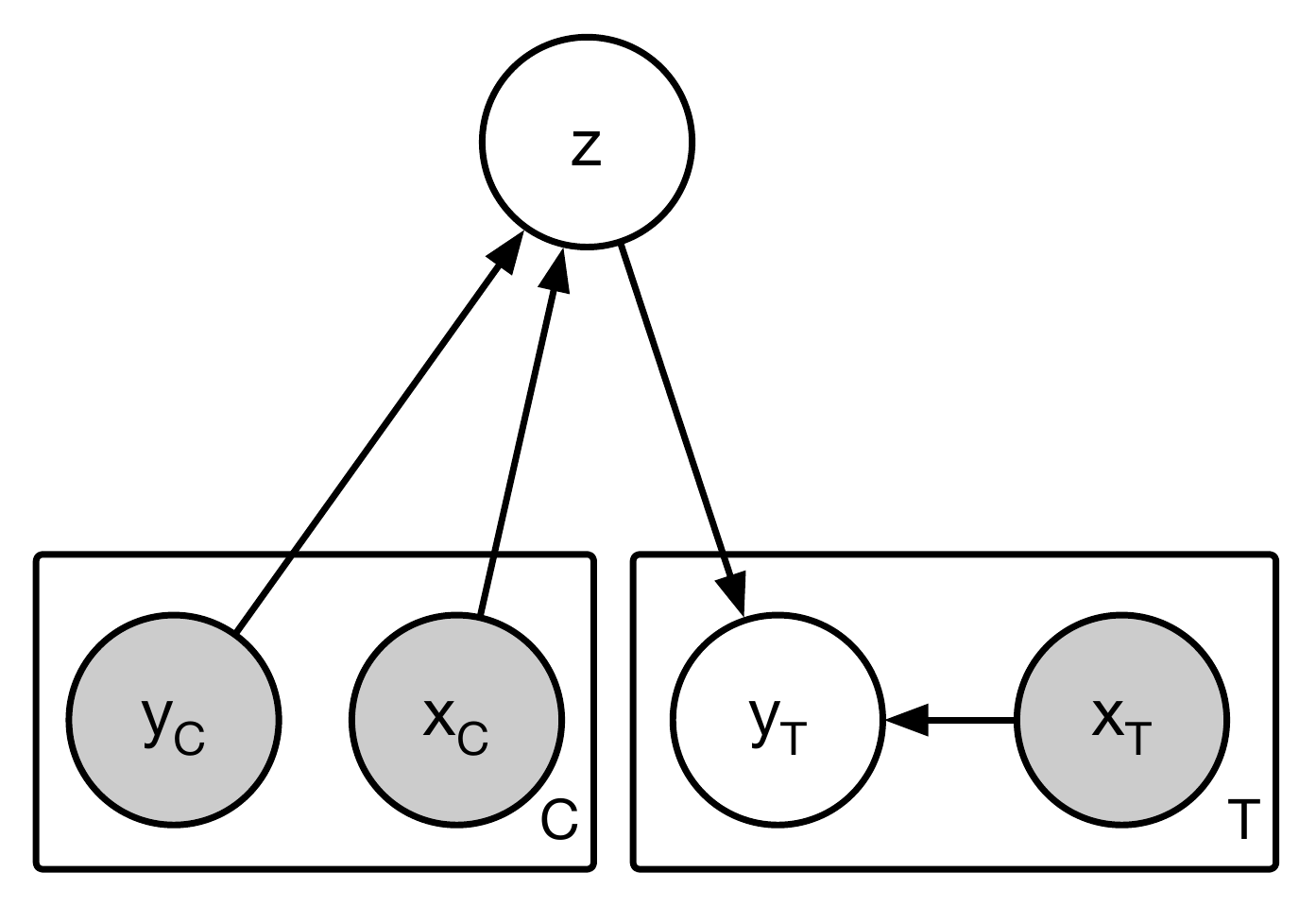}
        \caption{Graphical model}
        \label{fig:np0}
    \end{subfigure}
     ~ 
     \hspace{0.08\textwidth}
    \begin{subfigure}[b]{0.50\textwidth}
        \includegraphics[width=\textwidth]{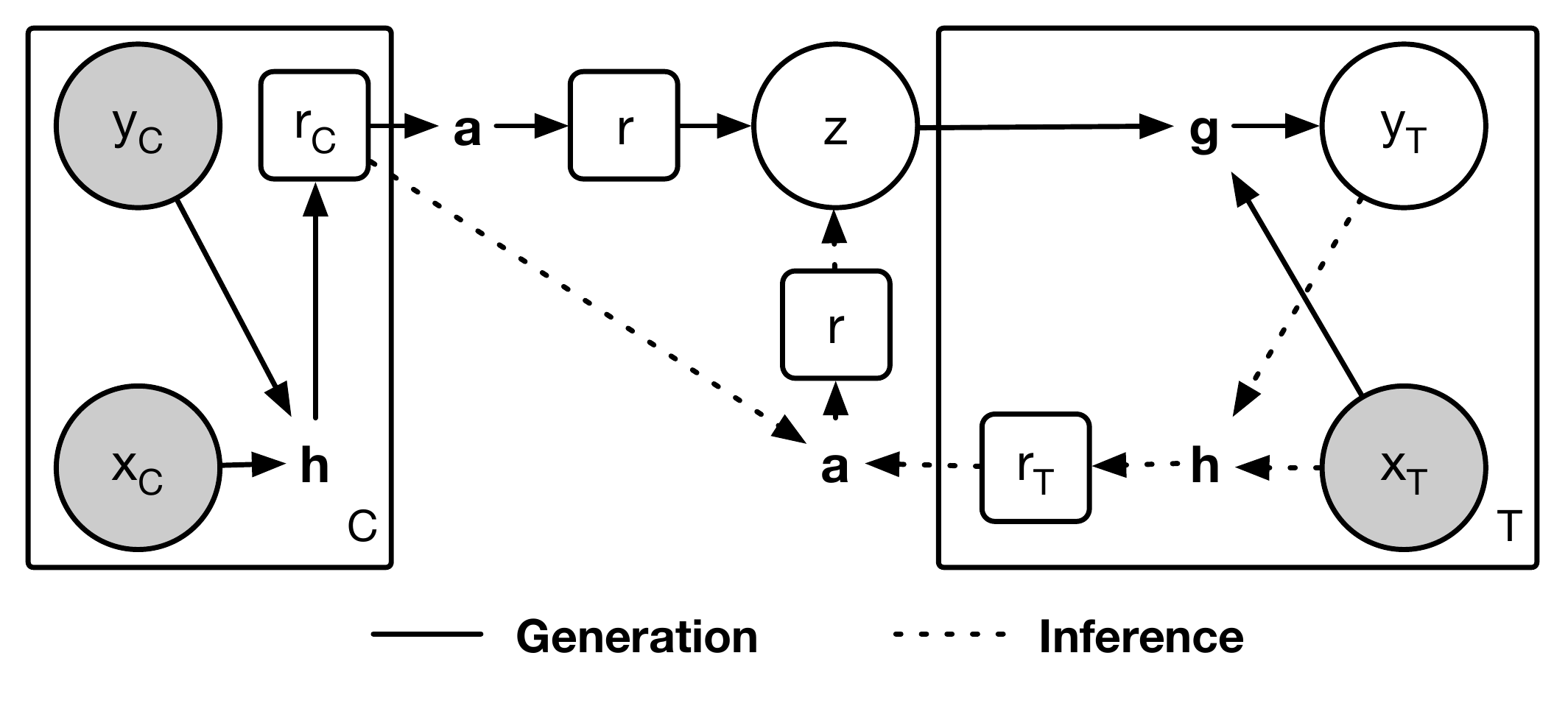}
        \caption{Computational diagram}
        \label{fig:np_comp}
    \end{subfigure}
    \caption{\textbf{Neural process model.} \textbf{(a)} Graphical model of a neural process. $x$ and $y$ correspond to the data where $y=f(x)$. $C$ and $T$ are the number of context points and target points respectively and $z$ is the global latent variable. A grey background indicates that the variable is observed. \textbf{(b)} Diagram of our neural process implementation. Variables in circles correspond to the variables of the graphical model in (a), variables in square boxes to the intermediate representations of NPs and unbound, bold letters to the following computation modules: $h$ - encoder, $a$ - aggregator and $g$ - decoder. In our implementation $h$ and $g$ correspond to neural networks and $a$ to the mean function. The continuous lines depict the generative process, the dotted lines the inference.}
    \label{fig:NP_models}
\end{figure*}

Given a collection of joint distributions $\rho_{x_{1:n}}$ we can derive two necessary conditions to be able to define a stochastic process $F$ such that $\rho_{x_{1:n}}$ is the marginal distribution of $(F(x_1),\ldots,F(x_n))$, for each finite sequence $x_{1:n}$. These conditions are: (finite) exchangeability and consistency. As stated by the Kolmogorov Extension Theorem~\citep{oksendal2003stochastic} these conditions are sufficient to define a stochastic process.

\textbf{Exchangeability} This condition requires the joint distributions to be invariant to permutations of the elements in $x_{1:n}$.  More precisely, for each finite $n$, if $\pi$ is a permutation of $\{1,\ldots,n\}$, then: 
\begin{align}
    &\rho_{x_{1:n}}(y_{1:n}) := \rho_{x_1,\ldots,x_n}(y_1,\ldots,y_n) \label{exch} \\ =& 
    \rho_{x_{\pi(1)},\ldots,x_{\pi(n)}}(y_{\pi(1)},\ldots,y_{\pi(n)}) =: \rho_{\pi(x_{1:n})}(\pi(y_{1:n}))   \nonumber 
\end{align}
where $\pi(x_{1:n}):=(x_{\pi(1)},\ldots,x_{\pi(n)})$ and $\pi(y_{1:n}):=(y_{\pi(1)},\ldots,y_{\pi(n)})$.  

\textbf{Consistency} If we marginalise out a part of the sequence the resulting marginal distribution is the same as that defined on the original sequence. More precisely, if $1\le m\le n$, then:
\begin{align}
    \rho_{x_{1:m}}(y_{1:m}) 
    =& \int \rho_{x_{1:n}}(y_{1:n}) dy_{m+1:n}.
    \label{const} 
\end{align}
Take, for example, three different sequences $x_{1:n}, \pi(x_{1:n})$ and $x_{1:m}$ as well as their corresponding joint distributions $\rho_{x_{1:n}}, \rho_{\pi(x_{1:n})}$ and $\rho_{x_{1:m}}$. In order for these joint distributions to all be marginals of some higher-dimensional distribution given by the stochastic process $F$, they have to satisfy equations~\ref{exch} and~\ref{const} above.

Given a particular instantiation of the stochastic process $f$ the joint distribution is defined as:
\begin{align}
    \rho_{x_{1:n}}(y_{1:n}) &= \int p(f) p(y_{1:n}|f,x_{1:n}) df. \label{eq:joint}
\end{align}
Here $p$ denotes the abstract probability distribution over all random quantities. Instead of $Y_i=F(x_i)$, we add some observation noise $Y_i \sim \mathcal{N}(F(x_i),\sigma^2)$ and define $p$ as:
\begin{align}
    p(y_{1:n}|f,x_{1:n}) &= \prod_{i=1}^n \mathcal{N}(y_i| f(x_i),\sigma^2).
\end{align}
Inserting this into equation~\ref{eq:joint} the stochastic process is specified by:
\begin{align}
    \rho_{x_{1:n}}(y_{1:n}) &= \int p(f) \prod_{i=1}^n \mathcal{N}(y_i| f(x_i),\sigma^2) df. \label{eq:condiid}
\end{align}

In other words, exchangeability and consistency of the collection of joint distributions $\{\rho_{x_{1:n}}\}$ impliy the existence of a stochastic process $F$ such that the observations $Y_{1:n}$ become iid conditional upon $F$.  This essentially corresponds to a conditional version of de Finetti's Theorem that anchors much of Bayesian nonparametrics \cite{de1937prevision}.
In order to represent a stochastic process using a NP, we will approximate it with a neural network, and assume that $F$ can be parameterised by a high-dimensional random vector $z$, and write $F(x) = g(x,z)$ for some fixed and learnable function $g$ (i.e.\ the randomness in $F$ is due to that of $z$). The generative model (Figure~\ref{fig:np0}) then follows from \eqref{eq:condiid}:
\begin{align}
    p(z,y_{1:n}|x_{1:n}) = p(z) \prod_{i=1}^n \mathcal{N}(y_i| g(x_i,z),\sigma^2)
\end{align}
where, following ideas of variational auto-encoders, we assume $p(z)$ is a multivariate standard normal, and $g(x_i,z)$ is a neural network which captures the complexities of the model. 

To learn such a distribution over random functions, rather than a single function, it is essential to train the system using multiple datasets concurrently, with each dataset being a sequence of inputs $x_{1:n}$ and outputs $y_{1:n}$, so that we can learn the variability of the random function from the variability of the datasets (see section~\ref{data}). 

Since the decoder $g$ is non-linear, we can use amortised variational inference to learn it. Let $q(z|x_{1:n},y_{1:n})$ be a variational posterior of the latent variables $z$, parameterised by another neural network that is invariant to permutations of the sequences $x_{1:n}, y_{1:n}$. Then the evidence lower-bound (ELBO) is given by:
\begin{align}
    &\log p(y_{1:n}|x_{1:n}) \\
    \ge& \EE_{q(z|x_{1:n},y_{1:n})} \left[
    \sum_{i=1}^n \log p(y_i|z,x_i) + \log\frac{p(z)}{q(z|x_{1:n},y_{1:n})} 
    \right] \nonumber
\end{align}

In an alternative objective that better reflects the desired model behaviour at test time, we split the dataset into a context set, $x_{1:m},y_{1:m}$ and a target set $x_{m+1:n},y_{m+1:n}$, and model the conditional of the target given the context.  
This gives:
\begin{footnotesize}
\begin{equation}
\label{eq:elbo1}
\begin{split}
    &\log p(y_{m+1:n}|x_{1:n},y_{1:m}) \\
    \ge& \EE_{q(z|x_{1:n},y_{1:n})} \left[
    \sum_{i=m+1}^n \log p(y_i|z,x_i) + \log\frac{p(z|x_{1:m},y_{1:m})}{q(z|x_{1:n},y_{1:n})} 
    \right]
\end{split}
\end{equation}
\end{footnotesize}
%
Note that in the above the conditional prior $p(z|x_{1:m},y_{1:m})$ is intractable.  We can approximate it using the variational posterior $q(z|x_{1:m},y_{1:m})$, which gives,
\begin{footnotesize}
\begin{align}
    \label{final_elbo}
    &\log p(y_{m+1:n}|x_{1:n},y_{1:m}) \\
    \ge& \EE_{q(z|x_{1:n},y_{1:n})} \left[
    \sum_{i=m+1}^n \log p(y_i|z,x_i) + \log\frac{q(z|x_{1:m},y_{1:m})}{q(z|x_{1:n},y_{1:n})} 
    \right] \nonumber
\end{align}
\end{footnotesize}

\subsection{Distributions over functions}
\label{data}

A key motivation for NPs is the ability to represent a distribution over functions rather than a single function. 
In order to train such a model we need a training procedure that reflects this task.

More formally, to train a NP we form a dataset that consists of functions $f: X \to Y$ that are sampled from some underlying distribution $\mathcal{D}$. 
As an illustrating example consider a dataset consisting of functions $f_d(x) \sim \mathcal{GP}$ that have been generated using a Gaussian process with a fixed kernel. 
For each of the functions $f_d(x)$ our dataset contains a number of $(x, y)_i$ tuples  where $y_i = f_d(x_i)$. 
For training purposes we divide these points into a set of $n$ context points $C = \{(x, y)_i\}_{i=1}^n$ and a set of $n+m$ target points which consists of all points in $C$ as well as $m$ additional unobserved points $T = \{(x, y)_i\}_{i=1}^{n+m}$.
During testing the model is presented with some context $C$ and has to predict the target values $y_T = f(x_T)$ at target positions $x_T$.

In order to be able to predict accurately across the entire dataset a model needs to learn a distribution that covers all of the functions observed in training and be able to take into account the context data at test time.

\begin{figure*}
    \centering
    \begin{subfigure}[b]{0.17\textwidth}
        \includegraphics[width=\textwidth]{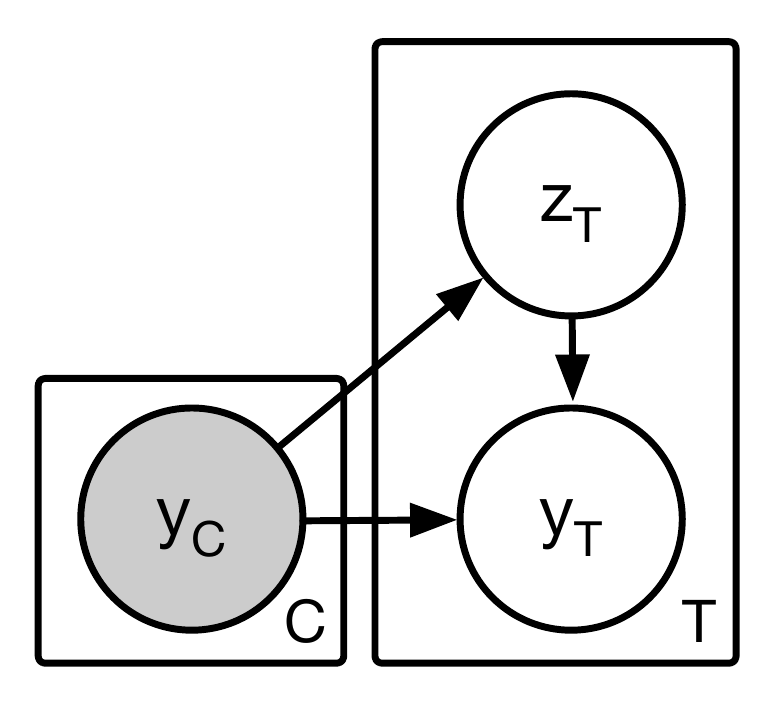}
        \caption{Conditional VAE}
        \label{fig:cvae}
    \end{subfigure}
    ~ 
    \begin{subfigure}[b]{0.17\textwidth}
        \includegraphics[width=\textwidth]{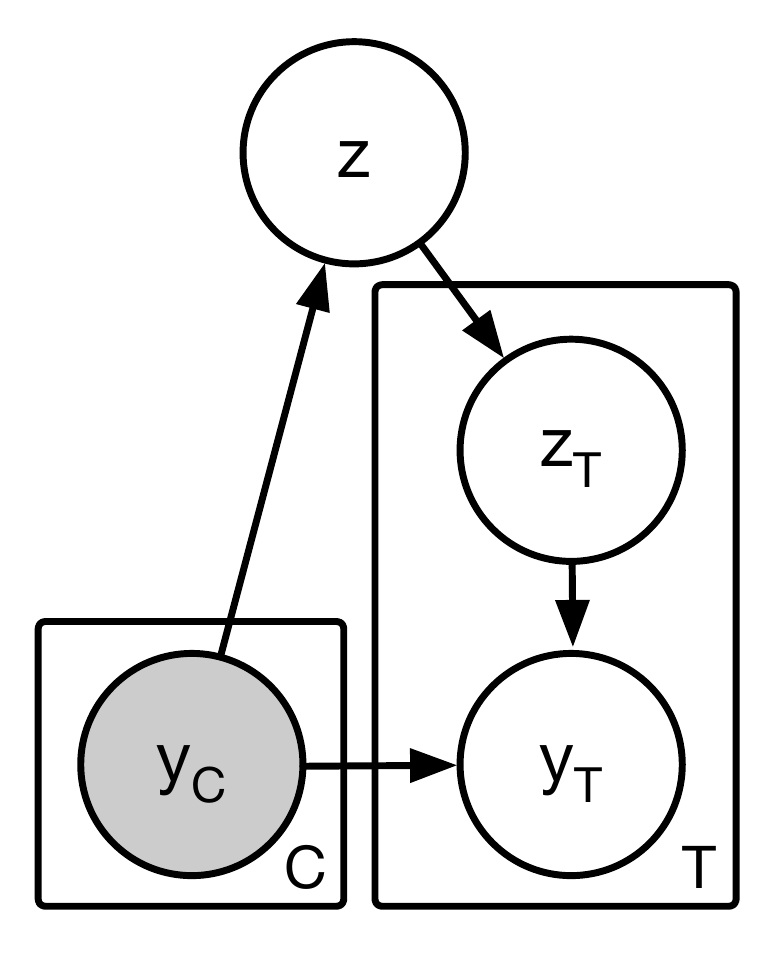}
        \caption{Neural statistician}
        \label{fig:ns}
    \end{subfigure}
    ~ 
    \begin{subfigure}[b]{0.29\textwidth}
        \includegraphics[width=\textwidth]{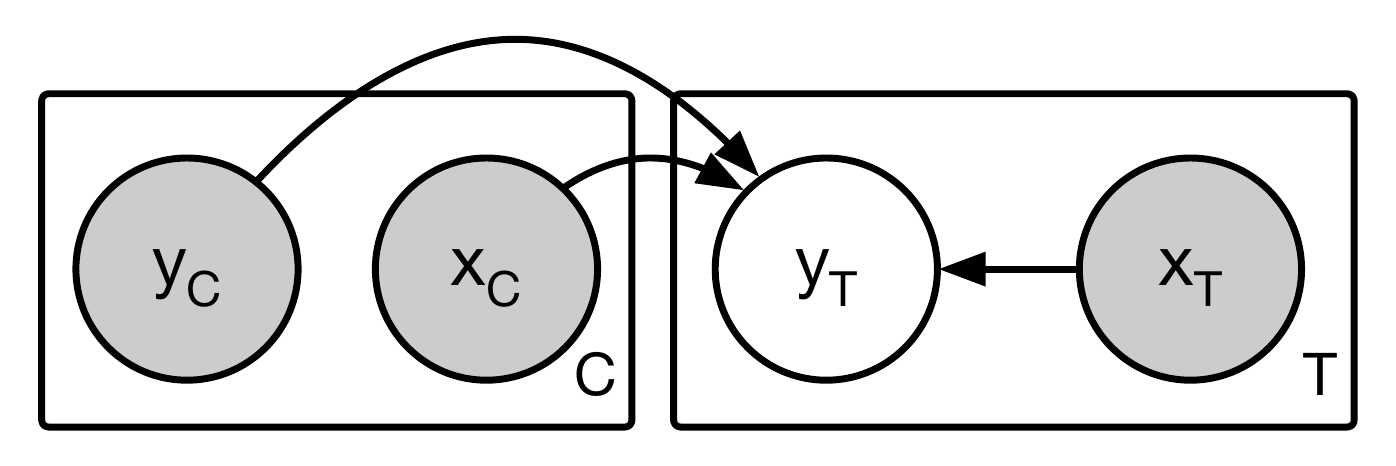}
        \caption{Conditional neural process}
        \label{fig:cnp}
    \end{subfigure}
     ~ 
    \begin{subfigure}[b]{0.29\textwidth}
        \includegraphics[width=\textwidth]{np_gm.pdf}
        \caption{Neural process}
        \label{fig:np}
    \end{subfigure}
    \caption{Graphical models of related models (a-c) and of the neural process (d). Gray shading indicates the variable is observed. $C$ stands for context variables and $T$ for target variables i.e. the variables to predict given $C$.}
    \label{fig:Models}
\end{figure*}

\subsection{Global latent variable}
\label{latent}

As mentioned above, neural processes include a latent variable $z$ that captures $F$. 
This latent variable is of particular interest because it captures the global uncertainty, which allows us to sample at a global level -- one function $f_d$ at a time,  rather than at a local output level -- one $y_i$ value for each $x_i$ at a time (independently of the remaining $y_T$).

In addition, since we are passing all of the context's information through this single variable we can formulate the model in a Bayesian framework.  
In the absence of context points $C$ the latent distribution $p(z)$ would correspond to a data specific prior the model has learned during training.
As we add observations the latent distribution encoded by the model amounts to the posterior $p(z|C)$ over the function given the context. 
On top of this, as shown in equation~\ref{final_elbo}, instead of using a zero-information prior $p(z)$, we condition the prior on the context. As such this prior is equivalent to a less informed posterior of the underlying function. 
This formulation makes it clear that the posterior given a subset of the context points will serve as the prior when additional context points are included. 
By using this setup, and training with different sizes of context, we encourage the learned model to be flexible with regards to the number and position of the context points.

\def\EE{\mathbb{E}}

\subsection{The Neural process model}
\label{npm}
In our implementation of NPs we accommodate for two additional desiderata: invariance to the order of context points and computational efficiency.
The resulting model can be boiled down to three core components (see Figure~\ref{fig:np_comp}):

\begin{itemize}
    \item An \textbf{encoder} $h$ from input space into representation space that takes in \emph{pairs} of $(x, y)_i$ context values and produces a representation $r_i = h((x, y)_i)$ for each of the pairs.
    We parameterise $h$ as a neural network.
    \item An \textbf{aggregator} $a$ that summarises the encoded inputs.
    We are interested in obtaining a single order-invariant global representation $r$ that parameterises the latent distribution $z \sim \mathcal{N}(\mu(r), I\sigma(r))$.
    The simplest operation that ensures order-invariance and works well in practice is the mean function  $r = a(r_i) =  \frac{1}{n}\sum_{i=1}^n r_i$. 
    Crucially, the aggregator reduces the runtime to $\mathcal{O}(n+m)$ where $n$ and $m$ are the number of context and target points respectively.
    \item A \textbf{conditional decoder} $g$ that takes as input the sampled global latent variable $z$ as well as the new target locations $x_T$ and outputs the predictions $\hat{y}_T$ for the corresponding values of $f(x_T) = y_T$.
\end{itemize}

\section{Related work}
\label{related}


\subsection{Conditional neural processes }
Neural Processes (NPs) are a generalisation of \textit{Conditional Neural Processes} (CNPs, \citet{garnelo2018conditional}).
CNPs share a large part of the motivation behind neural processes, but lack a latent variable that allows for global sampling (see Figure~\ref{fig:cnp} for a diagram of the model). As a result, CNPs are unable to produce different function samples for the same context data, which can be important if modelling this uncertainty is desirable. It is worth mentioning that the original CNP formulation did include experiments with a latent variable in addition to the deterministic connection. However, given the deterministic connections to the predicted variables, the role of the global latent variable is not clear. In contrast, NPs constitute a more clear-cut generalisation of the original deterministic CNP with stronger parallels to other latent variable models and approximate Bayesian methods. These parallels allow us to compare our model to a wide range of related research areas in the following sections.

Finally, NPs and CNPs themselves can be seen as generalizations of recently published generative query networks (GQN) which apply a similar training procedure to predict new viewpoints in 3D scenes given some context observations~\citep{eslami2018neural}. Consistent GQN (CGQN)  is an extension of GQN that focuses on generating consistent samples and is thus also closely related to NPs~\citep{ananyak_cgqn}.

\subsection{Gaussian processes}

We start by considering models that, like NPs, lie on the spectrum between neural networks (NNs) and Gaussian processes (GPs). 
Algorithms on the NN end of the spectrum fit a single function that they learn from a very large amount of data directly.
GPs on the other hand can represent a distribution over a family of functions, which is constrained by an assumption on the functional form of the covariance between two points.

Scattered across this spectrum, we can place recent research that has combined ideas from Bayesian non-parametrics with neural networks. 
Methods like~\citep{calandra2016manifold, huang2015scalable} remain fairly close to the GPs, but incorporate NNs to pre-process the input data. Deep GPs have some conceptual similarity to NNs as they stack GPs to obtain \emph{deep} models~\citep{damianou2013deep}. Approaches that are more similar to NNs include for example neural networks whose weights are sampled using a GPs~\citep{wilson2011gaussian} or networks where each unit represents a different kernel~\citep{sun2018differentiable}. 

There are two models on this spectrum that are closely related to NPs: matching networks (MN, \citet{vinyals2016matching}) and deep kernel learning (DKL, \citet{wilson2016deep}). As with NPs both use NNs to extract representations from the data, but while NPs learn the `kernel' to compare data points implicitly these other two models pass the representation to an explicit distance kernel. 
MNs use this kernel to measure the similarity between contexts and targets for few shot classification while the kernel in DKL is used to parametrise a GP.  
Because of this explicit kernel the computational complexity of MNs and DKL would be quadratic and cubic instead of $\mathcal{O}(n+m)$ like it is for NPs. 
To overcome this computational complexity DKL replace a standard GP with a kernel approximation given by a KISS GP~\citep{wilson2015kernel}, while prototypical networks~\citep{snell2017prototypical} are introduced as a more light-weight version of MNs that also scale with $\mathcal{O}(n+m)$.

Finally concurrent work by Ma et al introduces variational implicit processes, which share large part of the motivation of NPs but are implemented as GPs~\citep{ma2018variational}. In this context NPs can be interpreted as a neural implementation of an implicit stochastic process.



On this spectrum from NNs to GPs, neural processes remain closer to the neural end than most of the models mentioned above. 
By giving up on the explicit definition of a kernel NPs lose some of the mathematical guarantees of GPs, but trade this off for data-driven `priors' and computational efficiency. 
\begin{figure}
    \centering
    \includegraphics[width=\columnwidth]{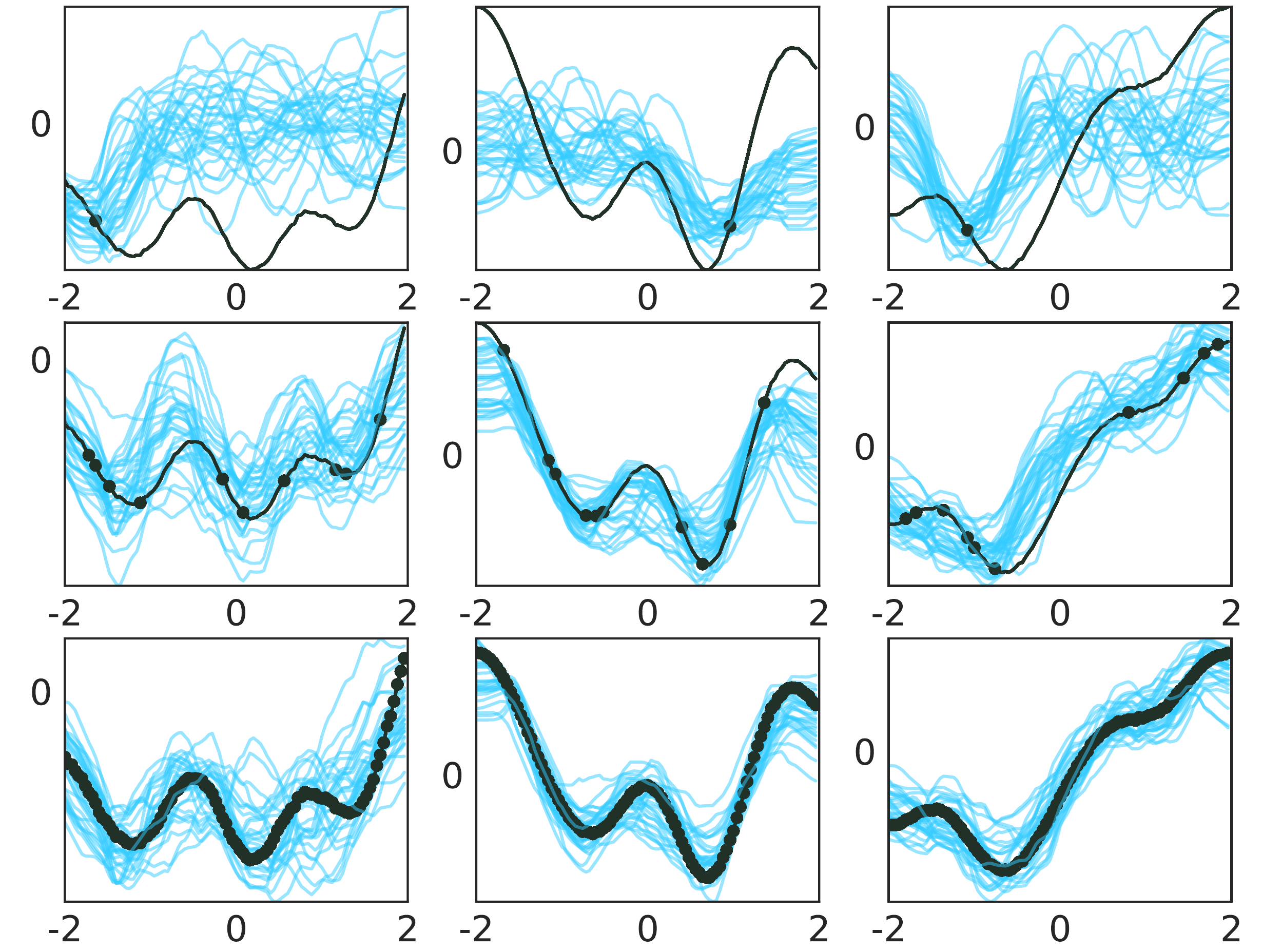}
        \caption{\textbf{1-D function regression.} The plots show samples of curves conditioned on an increasing number of context points (1, 10 and 100 in each row respectively). The true underlying curve is shown in black and the context points as black circles. Each column corresponds to a different example. With only one observation the variance away from that context point is high between the sampled curves. As the number of context points increases the sampled curves increasingly resemble the ground truth curve and the overall variance is reduced.}
    \label{fig:1dreg}
\end{figure}
\subsection{Meta-learning}


In contemporary meta-learning vocabulary, NPs and GPs can be seen to be methods for `few-shot function estimation'. In this section we compare with related models that can be used to the same end. A prominent example is matching networks~\cite{vinyals2016matching}, but there is a large literature of similar models for classification \citep{koch2015siamese, santoro2016one}, reinforcement learning~\citep{wang2016learning}, parameter update~\citep{finn2017model, finn2018probabilistic}, natural language processing~\citep{bowman2015generating} and program induction~\citep{devlin2017neural}. Related are generative meta-learning approaches that carry out few-shot estimation of the data densities~\citep{van2016conditional, reed2017few, bornschein2017variational, rezende2016one}.

Meta-learning models share the fundamental motivations of NPs as they shift workload from training time to test time. 
NPs can therefore be described as meta-learning algorithms for few-shot function regression, although as shown in~\citet{garnelo2018conditional} they can also be applied to few-shot learning tasks beyond regression. 

%

\subsection{Bayesian methods}
The link between meta-learning methods and other research areas, like GPs and Bayesian methods, is not always evident. 
Interestingly, recent work by ~\citet{grant2018recasting} out the relation between model agnostic meta learning (MAML, \citet{finn2017model}) and hierarchical Bayesian inference. 
In this work the meta-learning properties of MAML are formulated as a result of task-specific variables that are conditionally independent given a higher level variable. 
This hierarchical description can be rewritten as a probabilistic inference problem and the resulting marginal likelihood $p(y|C)$ matches the original MAML objective. 
The parallels between NPs and hierarchical Bayesian methods are similarly straightforward. Given the graphical model in Figure~\ref{fig:np} we can write out the conditional marginal likelihood as a hierarchical inference problem:
\begin{align}
\label{eq:elbo2}
    \log p(y_t|C, x_t) = \log \int  p(y_t|z, x_t)  p(z|C) dz
\end{align}
Another interesting area that connects Bayesian methods and NNs are Bayesian neural networks~\citep{gal2016dropout, blundell2015weight, louizos2017bayesian, louizos2017multiplicative}. These models learn distributions over the network weights and use the posterior of these weights to estimate the values of $y_T$ given $y_C$. In this context NPs can be thought of as amortised version of Bayesian DL. 



\subsection{Conditional latent variable models}

We have covered algorithms that are conceptually similar to NPs and algorithms that carry out similar tasks to NPs. In this section we look at models of the same family as NPs: conditional latent variable models.
Such models (Figure~\ref{fig:cvae}) learn the conditional distribution $p(y_{T}|y_{C}, z)$ where $z$ is a latent variable that can be sampled to generate different predictions. Training this type of directed graphical model is intractable and as with variational autoencoders (VAEs, \citet{rezende2014stochastic, kingma2013auto}), conditional variational autoencoders (CVAEs, \citet{sohn2015learning}) approximate the objective function using the variational lower bound on the log likelihood:
\begin{equation}
\label{eq:elbo3}
\begin{split}
    \log p(y_t|y_c) \geq \mathbb{E}_{q(z_{T}|y_c, y_t)}\Bigg[& \log  p(y_t|z_{T}, y_c) \\
    & + \log \frac{p(z_{T}|y_c)}{q(z_{T}|y_c, y_t)}\Bigg]
\end{split}
\end{equation}
We refer to the latent variable of CVAEs $z_{T}$ as a local latent variable in order to distinguish from global latent variables that are present in the models later on. 
We call this latent variable local as it is sampled anew for each of the output predictions $y_{T, i}$. This is in contrast to a global latent variable that is only sampled once and used to predict multiple values of $y_t$.
In the CVAE, conditioning on the context is done by adding the dependence both in the prior $p(z_{T}|y_c)$ and decoder $p(y|z_{T},y_c)$ so they can be considered as deterministic functions of the context. 

CVAEs have been extended in a number of ways for example by adding attention~\citep{rezende2016one}.
Another related extension is generative matching networks (GMNs, \citet{bartunov2016fast}), where the conditioning input is pre-processed in a way that is similar to the matching networks model.

A more complex version of the CVAE that is very relevant in this context is the neural statistician (NS, \citet{edwards2016towards}). 
Similar to the neural process, the neural statistician contains a global latent variable $z$ that captures global uncertainty (see Figure~\ref{fig:ns}).
A crucial difference is that while NPs represent the distribution over functions, NS represents the distribution over sets.
Since NS does not contain a corresponding $x$ value for each $y$ value, it does not capture a pair-wise relation like GPs and NPs, but rather a general distribution of the $y$ values. 
Rather than generating different $y$ values by querying the model with different $x$ values, NS generates different $y$ values by sampling an additional local hidden variable $z_{T}$.

\begin{figure*}[h]
    \centering
    \begin{subfigure}[b]{0.3\textwidth}
    \raisebox{.2\textwidth}{%
        \includegraphics[width=\textwidth]{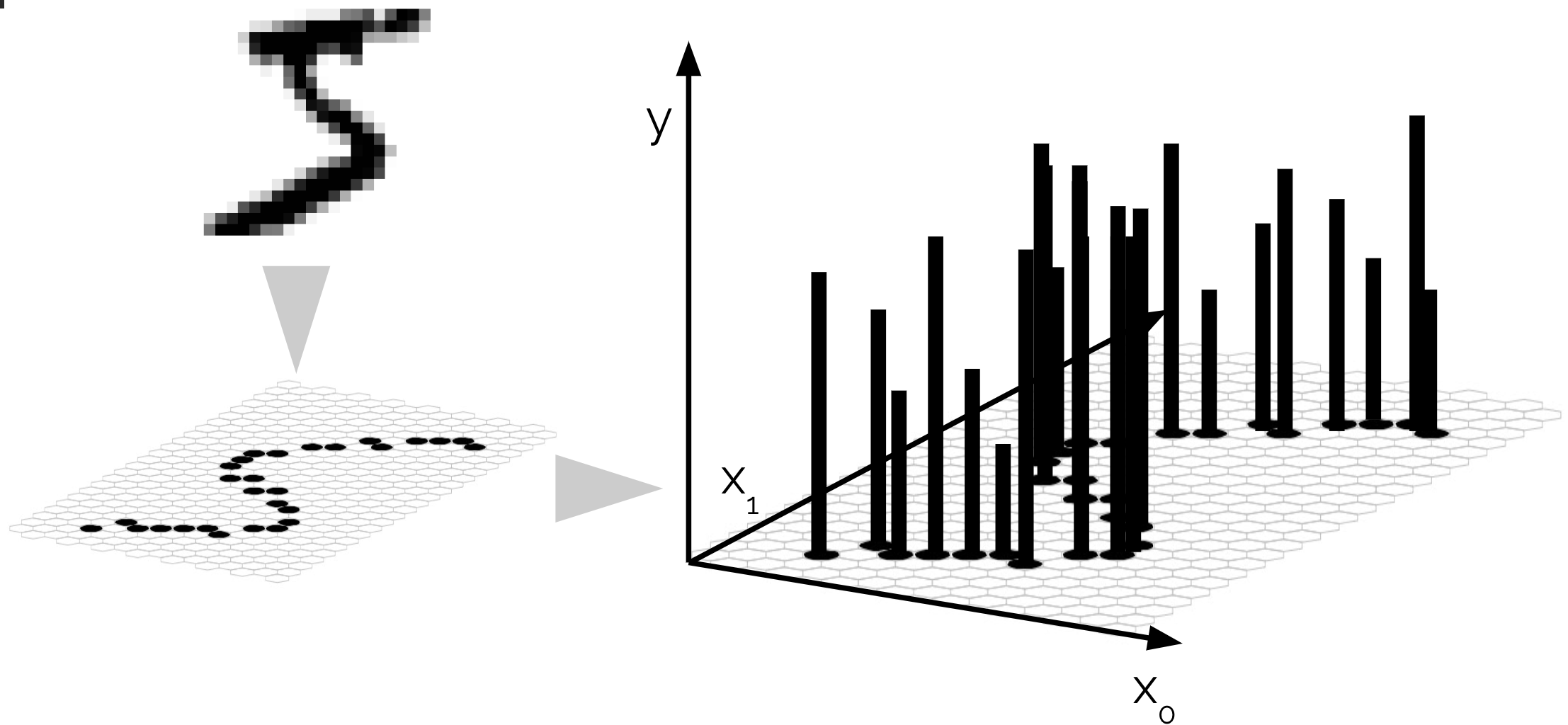}}
        \label{fig:overview}
    \end{subfigure}
    ~ 
    \begin{subfigure}[b]{0.3\textwidth}
        \includegraphics[width=\textwidth]{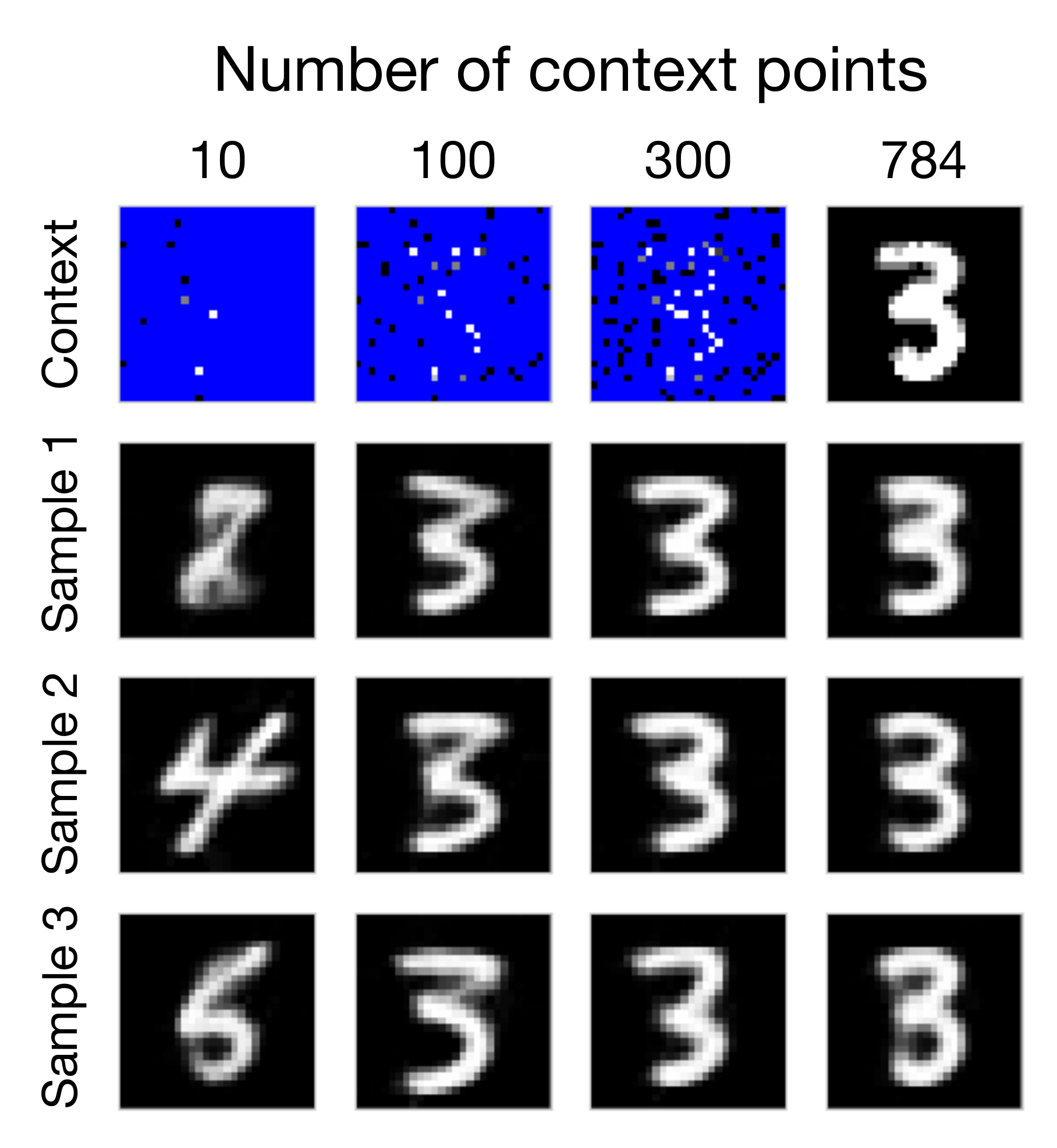}
        \label{fig:mnist}
    \end{subfigure}
    ~ 
    \begin{subfigure}[b]{0.3\textwidth}
        \includegraphics[width=\textwidth]{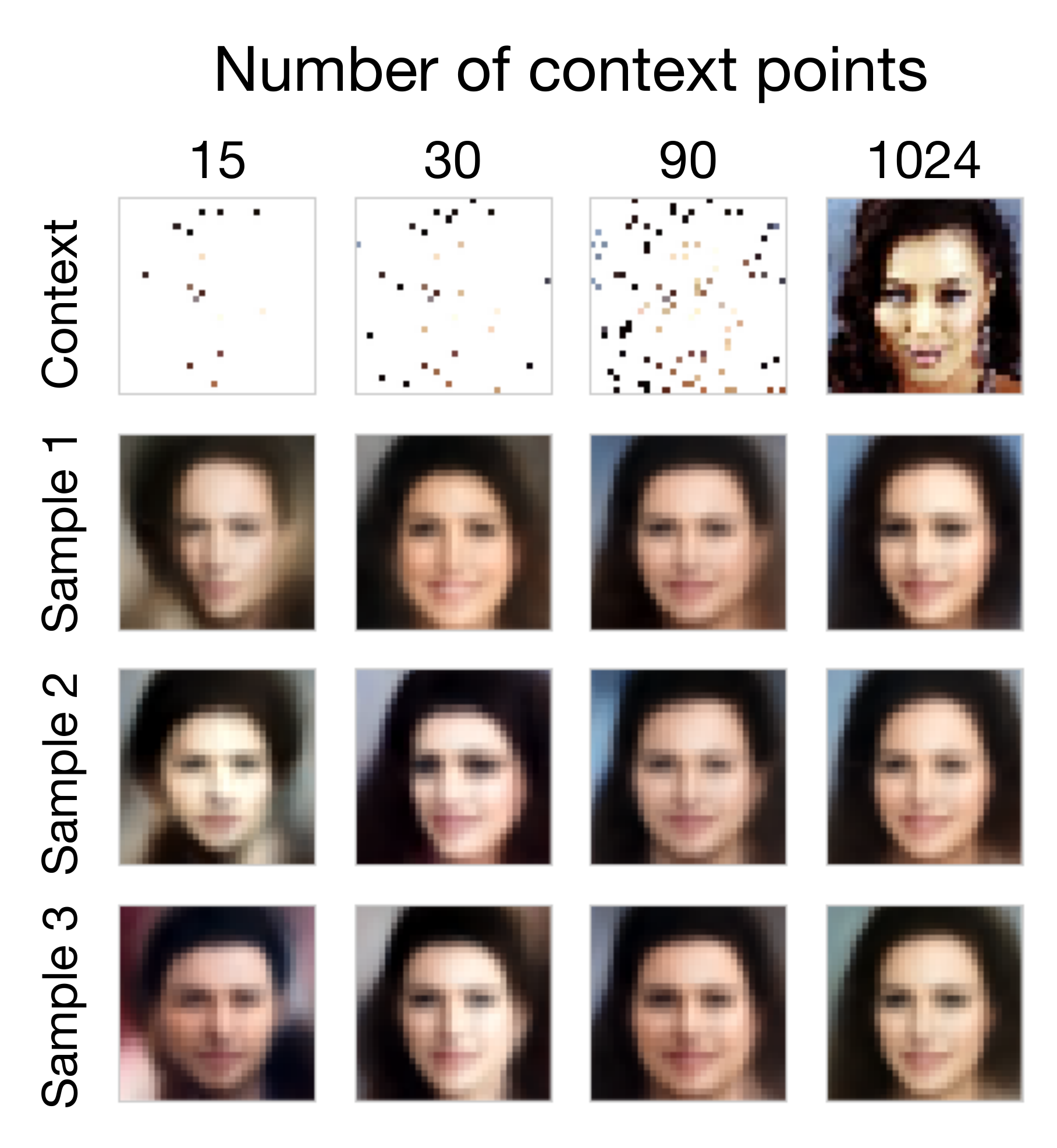}
        \label{fig:celeba}
    \end{subfigure}
    \caption{\textbf{Pixel-wise regression on MNIST and CelebA} The diagram on the left visualises how pixel-wise image completion can be framed as a 2-D regression task where f(pixel coordinates) = pixel brightness. The figures to the right of the diagram show the results on image completion for MNIST and CelebA. The images on the top correspond to the context points provided to the model. For better clarity the unobserved pixels have been coloured blue for the MNIST images and white for CelebA. Each of the rows corresponds to a different sample given the context points. As the number of context points increases the predicted pixels get closer to the underlying ones and the variance across samples decreases.}
    \label{2dreg}
\end{figure*}

\normalsize

The ELBO of the NS reflects the hierarchical nature of the model with a double expectation over the local and the global variable. If we leave out the local latent variable for a more direct comparison to NPs the ELBO becomes:
\begin{equation}
\begin{split}
    \log p(y_t, y_c) \geq   \mathbb{E}_{q(z|y_c, y_t)}\Bigg[ & \log p(y_t|z)\\
    & + \log \frac{p(z)}{q(z|y_c, y_t)}\Bigg]
\end{split}
\end{equation}
Notably the prior $p(z)$ of NS is not conditioned on the context. The prior of NPs on the other hand is conditional (equation~\ref{final_elbo}), which brings the training objective closer to the way the model is used at test time.
Another variant of the ELBO is presented in the variational homoencoder~\citep{hewitt2018variational}, a model that is very similar to the neural statistician but uses a separate subset of data points for the context and predictions. 

As reflected in Figure~\ref{fig:Models} the main difference between NPs and the conditional latent variable models is the lack of an $x$ variable that allows for targeted sampling of the latent distribution. This change, despite seeming small, drastically changes the range of applications. 
Targeted sampling, for example allows for generation and completion tasks (e.g. the image completion tasks) or the addition of some downstream task (like using an NP for reinforcement learning).
It is worth mentioning that all of the conditional latent variable models have also been applied to few-shot classification problems, where the data space generally consists of input tuples $(x, y)$, rather than just single outputs $y$. 
The models are able to carry this out by framing classification either as a comparison between the log likelihoods of the different classes or by looking at the KL between the different posteriors, thereby overcoming the need of working with data tuples.

\section{Results}
\label{results}

\subsection{1-D function regression}

In order to test whether neural processes indeed learn to model distributions over functions we first apply them to a 1-D function regression task. 
The functions for this experiment are generated using a GP with varying kernel parameters for each function. 
At every training step we sample a set of values for the Gaussian kernel of a GP and use those to sample a function $f_D(x)$. 
A random number of the $(x, y)_C$ pairs are passed into the decoder of the NP as context points. 
We pick additional unobserved pairs $(x, y)_U$ which we combine with the observed context points $(x, y)_C$ as targets and feed $x_T$ to the decoder that returns its estimate $\hat{y_T}$ of the underlying value of $y_T$.

Some sample curves are shown in Figure~\ref{fig:1dreg}. For the same underlying ground truth curve (black line) we run the neural process using varying numbers of context points and generate several samples for each run (light-blue lines). 
As evidenced by the results the model has learned some key properties of the 1-D curves from the data such as continuity and the general shape of functions sampled from a GP with a Gaussian kernel.
When provided with only one context point the model generates curves that fluctuate around 0, the prior of the data-generating GP.
Crucially, these curves go through or near the observed context point and display a higher variance in regions where no observations are present.
As the number of context points increases this uncertainty is reduced and the model's predictions better match the underlying ground truth.
Given that this is a neural approximation the curves will sometimes only approach the observations points as opposed to go through them as it is the case for GPs.  
On the other hand once the model is trained it can regress more than just one data set i.e. it will produce sensible results for curves generated using any kernel parameters observed during training.

\subsection{2-D function regression}
One of the benefits of neural processes is their functional flexibility as they can learn non-trivial `kernels' from the data directly. In order to test this we apply NPs to a more complex regression problem. We carry out image completion as a regression task, where we provide some of the pixels as context and do pixel-wise prediction over the entire image. In this formulation the $x_i$ values would correspond to the Cartesian coordinates of each pixel and the $y_i$ values to the pixel intensity (see Figure~\ref{2dreg} for an explanation of this). It is important to point out that we choose images as our dataset because they constitute a complex 2-D function and they are easy to evaluate visually. 
It is important to point out that NPs, as such, have not been designed for image generation like other specialised generative models.

We train separate models on the MNIST~\citep{lecun1998gradient} and the CelebA~\citep{liu2015faceattributes} datasets.
As shown in Figure~\ref{2dreg} the model performs well on both tasks. 
In the case of the MNIST digits the uncertainty is reflected in the variability of the generated digit. 
Given only a few context points more than just one digit can fit the observations and as a result the model produces different digits when sampled several times. 
As the number of context points increases the set of possible digits is reduced and the model produces the same digit, albeit with structural modifications that become smaller as the number of context points increases.

The same holds for the CelebA dataset. In this case, when provided limited context the model samples from a wider range of possible faces and as it observes more context points it converges towards very similar looking faces. We do not expect the model to reconstruct the target image perfectly even when all the pixels are provided as context, since the latent variable $z$ constitutes a strong bottleneck. This can be seen in the final column of the figure where the predicted images are not only not identical to the ground truth but also vary between themselves. The latter is likely a cause of the latent variance which has been clipped to a small value to avoid collapsing, so even when no uncertainty is present we can generate different samples from $p(z|C)$. 

\begin{figure*}
\begin{center}
\centerline{\includegraphics[width=\columnwidth*2]{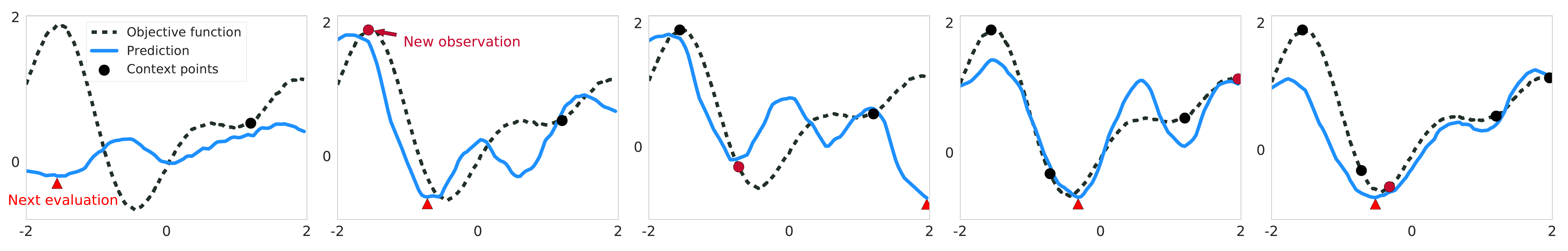}}
\caption{\textbf{Thompson sampling with neural processes on a 1-D objective function.} The plots show the optimisation process
         over five iterations. Each prediction function (blue) is drawn by sampling a latent variable conditioned on an incresing number of context points (black circles). The underlying ground truth function is depicted as a black dotted line. The red triangle indicates the next evaluation point which corresponds to the minimum value of the sampled NP curve. The red circle in the following iteration corresponds to this evaluation point with its underlying ground truth value that serves as a new context point to the NP. }
\label{fig:bo_v1}
\end{center}
\end{figure*}

\subsection{Black-box optimisation with Thompson sampling}

To showcase the utility of sampling entire consistent trajectories we apply neural processes to Bayesian optimisation on 1-D function using Thompson sampling \cite{thompson1933likelihood}. Thompson sampling (also known as randomised probability matching) is an approach to tackle the exploration-exploitation dilemma by maintaining a posterior distribution over model parameters. A decision is taken by drawing a sample of model parameters and acting greedily under the resulting policy. The posterior distribution is then updated and the process is repeated. Despite its simplicity, Thompson sampling has been shown to be highly effective both empirically and in theory. It is commonly applied to black box optimisation and multi-armed bandit problems \citep[e.g.][]{agrawal2012analysis, shahriari2016taking}. 

Neural processes lend themselves naturally to Thompson sampling by instead drawing a function over the space of interest, finding its minimum and adding the observed outcome to the context set for the next iteration. As shown in in Figure~\ref{fig:1dreg}, function draws show high variance when only few observations are available, modelling uncertainty in a similar way to draws from a posterior over parameters given a small data set. An example of this procedure for neural processes on a 1-D objective function is shown in Figure~\ref{fig:bo_v1}.

We report the average number of steps required by an NP to reach the global minimum of a function generated from a GP prior in Table~\ref{tab:bo_table}.
For an easier comparison the values are normalised by the amount of steps required when doing optimisation using random search. 
On average, NPs take four times fewer iterations than random search on this task.
An upper bound on performance is given by a Gaussian process with the same kernel than the GP that generated the function to be optimised. NPs do not reach this optimal performance, as their samples are more noisy than those of a GP, but are faster to evaluate since merely a forward pass through the network is needed. This difference in computational speed is bound to get more notable as the dimensionality of the problem and the number of necessary function evaluations increases.

\begin{table}[H]
\begin{center}
 \begin{tabular}{ c | c | c} 
 \textbf{Neural process} & \textbf{Gaussian process} & \textbf{Random Search} \\  [1ex] 
 \hline \hline
 0.26 & 0.14 & 1.00 \\ [1ex] 
\end{tabular}
\end{center}
\caption{\textbf{Bayesian Optimisation using Thompson sampling.} Average number of optimisation steps needed to reach the global minimum of a 1-D function generated by a Gaussian process. The values are normalised by the number of steps taken using random search. The performance of the Gaussian process with the correct kernel constitutes an upper bound on performance.}
\label{tab:bo_table}
\end{table}

\begin{table*}[h]
\small
\footnotesize
\centering
\caption{Results on the wheel bandit problem for increasing values of $\delta$. Shown are mean and standard errors for both cumulative and simple regret (a measure of the quality of the final policy) over 100 trials. Results normalised wrt. the performance of a uniform agent.}
\begin{tabular}{lcccccc@{}}
    \toprule
    {\bf $\delta$}  &
    {\bf 0.5} & {\bf 0.7} & {\bf 0.9} & {\bf 0.95} & {\bf 0.99} \\
    \midrule
    \textbf{\textit{Cumulative regret}} &  &  &  &  &  \\
    \\
    Uniform		                                    & 100.00 {\tiny $\pm$ 0.08 } & 100.00 {\tiny $\pm$ 0.09 } & 100.00 {\tiny $\pm$ 0.25 } & 100.00 {\tiny $\pm$ 0.37 } & 100.00 {\tiny $\pm$ 0.78 }\\
    LinGreedy ($\epsilon = 0.0$)                    & 65.89 {\tiny $\pm$ 4.90 } & 71.71 {\tiny $\pm$ 4.31 } & 108.86 {\tiny $\pm$ 3.10 } & 102.80 {\tiny $\pm$ 3.06 } & 104.80 {\tiny $\pm$ 0.91 }\\

    Dropout                                         & 7.89 {\tiny $\pm$ 1.51 } & 9.03 {\tiny $\pm$ 2.58 } &  36.58 {\tiny $\pm$ 3.62 } &  63.12 {\tiny $\pm$ 4.26 } & 98.68 {\tiny $\pm$ 1.59 }\\

    LinGreedy ($\epsilon = 0.05$)	                & 7.86 {\tiny $\pm$ 0.27 } &  9.58 {\tiny $\pm$ 0.35 } & 19.42 {\tiny $\pm$ 0.78 } & 33.06 {\tiny $\pm$ 2.06 } & 74.17 {\tiny $\pm$ 1.63 }\\

    Bayes by Backprob  \cite{blundell2015weight}    & 1.37 {\tiny $\pm$ 0.07 } & 3.32 {\tiny $\pm$ 0.80 } &  34.42 {\tiny $\pm$ 5.50 } & 59.04 {\tiny $\pm$ 5.59 } & 97.38 {\tiny $\pm$ 2.66 }\\
    NeuralLinear                                    & \textbf{0.95} {\tiny $\pm$ 0.02 } &  \textbf{1.60} {\tiny $\pm$ 0.03 } &  4.65 {\tiny $\pm$ 0.18 } &  9.56 {\tiny $\pm$ 0.36 } & 49.63 {\tiny $\pm$ 2.41 }\\
    
    \\
    MAML \cite{finn2017model}                       & 2.95 {\tiny $\pm$ 0.12 } & 3.11 {\tiny $\pm$ 0.16 } & 4.84 {\tiny $\pm$ 0.22 } & 7.01 {\tiny $\pm$ 0.33 } & 22.93 {\tiny $\pm$ 1.57 }\\
    Neural Processes                                & 1.60 {\tiny $\pm$ 0.06 } & 1.75 {\tiny $\pm$ 0.05 } & \textbf{3.31} {\tiny $\pm$ 0.10 } & \textbf{5.71} {\tiny $\pm$ 0.24 } & \textbf{22.13} {\tiny $\pm$ 1.23 }\\
    \midrule
    \textbf{\textit{Simple regret}} &  &  &  &  &  \\
    \\
    Uniform		                                    & 100.00 {\tiny $\pm$ 0.45 } & 100.00 {\tiny $\pm$ 0.78 } & 100.00 {\tiny $\pm$ 1.18 } & 100.00 {\tiny $\pm$ 2.21 } & 100.00 {\tiny $\pm$ 4.21 }\\

    LinGreedy ($\epsilon = 0.0$)                    & 66.59 {\tiny $\pm$ 5.02 } & 73.06 {\tiny $\pm$ 4.55 } & 108.56 {\tiny $\pm$ 3.65 } & 105.01 {\tiny $\pm$ 3.59 } & 105.19 {\tiny $\pm$ 4.14 }\\
    
    Dropout                                         & 6.57 {\tiny $\pm$ 1.48 } & 6.37 {\tiny $\pm$ 2.53 } &  35.02 {\tiny $\pm$ 3.94 } &  59.45 {\tiny $\pm$ 4.74 } & 102.12 {\tiny $\pm$ 4.76 }\\
    
    LinGreedy ($\epsilon = 0.05$)	                & 5.53 {\tiny $\pm$ 0.19 } &  6.07 {\tiny $\pm$ 0.24 } & 8.49 {\tiny $\pm$ 0.47 } & 12.65 {\tiny $\pm$ 1.12 } & 57.62 {\tiny $\pm$ 3.57 }\\

    Bayes by Backprob  \cite{blundell2015weight}    & 0.60 {\tiny $\pm$ 0.09 } & 1.45 {\tiny $\pm$ 0.61 } & 27.03 {\tiny $\pm$ 6.19 } & 56.64 {\tiny $\pm$ 6.36 } & 102.96 {\tiny $\pm$ 5.93 }\\
    
    NeuralLinear                                    & \textbf{0.33} {\tiny $\pm$ 0.04 } &  \textbf{0.79} {\tiny $\pm$ 0.07 } & \textbf{2.17} {\tiny $\pm$ 0.14 } &  \textbf{4.08} {\tiny $\pm$ 0.20 } & 35.89 {\tiny $\pm$ 2.98 }\\
    
    \\
    MAML \cite{finn2017model}                       & 2.49 {\tiny $\pm$ 0.12 } & 3.00 {\tiny $\pm$ 0.35 } & 4.75 {\tiny $\pm$ 0.48 } & 7.10 {\tiny $\pm$ 0.77 } & 22.89 {\tiny $\pm$ 1.41 }\\
    Neural Processes                                & 1.04 {\tiny $\pm$ 0.06 } & 1.26 {\tiny $\pm$ 0.21 } & 2.90 {\tiny $\pm$ 0.35 } & 5.45 {\tiny $\pm$ 0.47 } & \textbf{21.45} {\tiny $\pm$ 1.3}\\
    \bottomrule 
\end{tabular}
\label{tab:bandit_wheel}
\end{table*}

\subsection{Contextual bandits}

\begin{figure}
\begin{center}
\centerline{\includegraphics[width=\columnwidth]{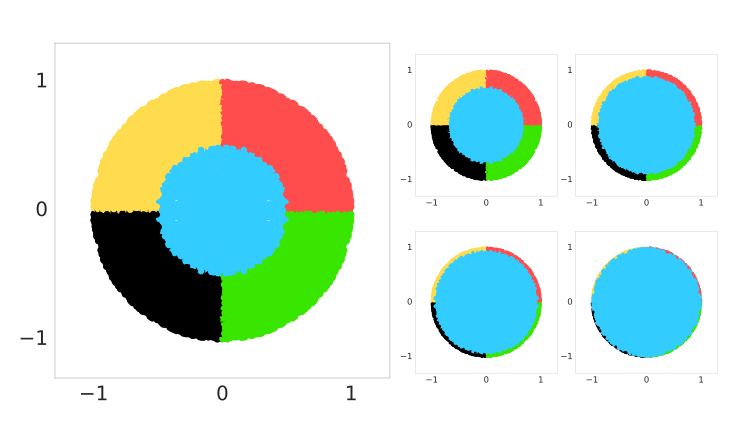}}
\caption{The wheel bandit problem with varying values of $\delta$.}
\label{fig:wheel_bandit}
\end{center}
\end{figure}

 Finally, we apply neural processes to the wheel bandit problem introduced in \citet{riquelme2018deep}, which constitutes a contextual bandit task on the unit circle with varying needs for exploration that can be smoothly parameterised. The problem can be summarised as follows (see Figure~\ref{fig:wheel_bandit} for clarity): a unit circle is divided into a low-reward region (blue area) and four high-reward regions (the other four coloured areas). The size of the low-reward region is defined by a scalar $\delta$. At every episode a different value for $\delta$ is selected. The agent is then provided with some coordinates $X = (X_1, X_2)$ within the circle and has to choose among $k=5$ arms depending on the area the coordinates fall into. If $||X|| \leq \delta$, the sample falls within the low-reward region (blue). In this case $k=1$ is the optimal action, as it provides a reward drawn from $r \sim \mathcal{}{N}(1.2, 0.01^2)$, while all other actions only return $r \sim \mathcal{N}(1.0, 0.01^2)$. If the sample falls within any of the four high-reward region ($||X|| > \delta$), the optimal arm will be any of the remaining four $k=2-5$, depending on the specific area. Pulling the optimal arm here results in a high reward $r \sim \mathcal{N}(50.0, 0.01^2)$, and as before all other arms receive $\mathcal{N}(1.0, 0.01^2)$ \emph{except} for arm $k=1$ which again returns $\mathcal{N}(1.2, 0.01^2)$.

 We compare our model to a large range of methods that can be used for Thompson sampling, taking results from \citet{riquelme2018deep}, who kindly agreed to share the experiment and evaluation code with us. Neural Processes can be applied to this problem by training on a distribution of tasks before applying the method. Since the methods described in \citet{riquelme2018deep} do not require such a pre-training phase, we also include Model-agnostic meta-learning (MAML, \citet{finn2017model}), a method relying on a similar pre-training phase, using code made available by the authors. For both NPs and MAML methods, we create a batch for pre-training by first sampling $M$ different wheel problems $\{\delta_i\}_{i=1}^M, \delta_i \sim \mathcal{U}(0, 1)$, followed by sampling tuples $\{(X, a, r)_j\}_{j=1}^{N}$ for context $X$, arm $a$ and associated reward $r$ for each $\delta_i$. We set $M=64, N=562$, using 512 context and 50 target points for Neural Processes, and an equal amount of data points for the meta- and inner-updates in MAML. Note that since gradient steps are necessary for MAML to adapt to data from each test problem, we reset the parameters after each evaluation run. This additional step is not necessary for neural processes. 
 
 Table \ref{tab:bandit_wheel} shows the quantitative evaluation on this task. We observe that Neural Processes are a highly competitive method, performing similar to MAML and the \textit{NeuralLinear} baseline in \citet{riquelme2018deep}, which is consistently among the best out of 20 algorithms compared.

\section{Discussion}
\label{discussion}

We introduce Neural processes, a family of models that combines the benefits of stochastic processes and neural networks. 
NPs learn to represent distributions over functions and make flexible predictions at test time conditioned on some context input. 
Instead of requiring a handcrafted kernel, NPs learn an implicit measure from the data directly.

We apply NPs to a range of regression tasks to showcase their flexibility. The goal of this paper is to introduce NPs and compare them to the currenly ongoing research. As such, the  tasks presented here are diverse but relatively low-dimensional. We leave it to future work to scale NPs up to higher dimensional problems that are likely to highlight the benefit of lower computational complexity and data driven representations.

\section*{Acknowledgements}
We would like to thank Tiago Ramalho, Oriol Vinyals, Adam Kosiorek, Irene Garnelo, Daniel Burgess, Kevin McKee and Claire McCoy for insightful discussions and being awesome people. We would also like to thank Carlos Riquelme, George Tucker and Jasper Snoek for providing the code to reproduce the results of their contextual bandits experiments (and, of course, also being awesome people).

\bibliography{references}
\bibliographystyle{icml2018}

\end{document}